# POMP: Probability-driven Meta-graph Prompter for LLMs in Low-resource Unsupervised Neural Machine Translation

Shilong Pan[1], Zhiliang Tian[1*], Liang Ding[2],
Zhen Huang[1], Zhihua Wen[1], Dongsheng Li[1*]
[1]College of Computer, National University of Defense Technology
[2]JD Explore Academy
{panshilong18, tianzhiliang, huangzhen, zhwen, dsli}@nudt.edu.cn
liangding.liam@gmail.com

## Abstract

Low-resource languages (LRLs) face challenges in supervised neural machine translation due to limited parallel data, prompting research into unsupervised methods. Unsupervised neural machine translation (UNMT) methods, including back-translation, transfer learning, and pivot-based translation, offer practical solutions for LRL translation, but they are hindered by issues like synthetic data noise, language bias, and error propagation, which can potentially be mitigated by Large Language Models (LLMs). LLMs have advanced NMT with in-context learning (ICL) and supervised fine-tuning methods, but insufficient training data results in poor performance in LRLs. We argue that LLMs can mitigate the linguistic noise with auxiliary languages to improve translations in LRLs. In this paper, we propose **Pr**Obability-driven **M**eta-graph **P**rompter (POMP), a novel approach employing a dynamic, sampling-based graph of multiple auxiliary languages to enhance LLMs' translation capabilities for LRLs. POMP involves constructing a directed acyclic meta-graph for each source language, from which we dynamically sample multiple paths to prompt LLMs to mitigate the linguistic noise and improve translations during training. We use the BLEURT metric to evaluate the translations and back-propagate rewards, estimated by scores, to update the probabilities of auxiliary languages in the paths. Our experiments show significant improvements in the translation quality of three LRLs, demonstrating the effectiveness of our approach.

## 1 Introduction

Training conventional neural machine translation (NMT) models usually in a supervised way, requires extensive parallel data (Liu et al., 2020; Raffel et al., 2020; Xue et al., 2021; Fan et al., 2021), therefore struggles in low-resource languages (LRLs) that have limited parallel datasets. Researchers extracting (Schwenk et al., 2021a,b) and annotating (Guzmán et al., 2019; Team et al., 2022; Goyal et al., 2022) parallel datasets of LRLs, which requires a substantial human effort. Therefore, the research community has explored diverse unsupervised methods, which operate without parallel data, to address the challenges.

---

*Corresponding Author

Unsupervised methods offer a practical and cost-effective alternative for translation in LRLs. These are primarily categorized into back-translation(Sennrich et al., 2016; Lample et al., 2018b), transfer learning(Chen et al., 2022; Li et al., 2022), and pivot-based translation(Kim et al., 2019; Currey and Heafield, 2019). (1) Back-translation employs NMT systems to translate the target language text into the source language, thus generating synthetic parallel data for training source-to-target translation models. However, this approach is constrained by the considerable noise in the synthetic parallel data (Epaliyana et al., 2021; Chauhan et al., 2022). (2) Transfer learning-based methods train models on high-resource languages and make inferences on LRLs. However, the language bias between training and testing languages weakens the effectiveness of this approach (Dabre et al., 2017). (3) Pivot-based translation leverages other languages as intermediary pivots between the source and target. This strategy facilitates the translation by first converting from the source to a pivot language, and then from the pivot to the target language. Nonetheless, the multi-hop translation process introduces an issue of noise propagation, which encompasses potential translation errors and language bias (Liu et al., 2019). Overall, these methods are impacted by noise: back-translation by synthetic data noise, transfer learning by language bias, and pivot-based translation by error propagation. We suggest Large Language Models (LLMs) can mitigate this linguistic noise in LRLs translation.

LLMs have greatly advanced NMT (Peng et al., 2023; Hendy et al., 2023; Zhu et al., 2023). Researchers have investigated in-context learning (ICL) (Brown et al., 2020; Dong et al., 2023) methods to leverage LLMs for NMT tasks (Vilar et al., 2023; Garcia et al., 2023; Zhang et al., 2023). ICL constructs the prompt with task exemplars and the input query in a template, enabling LLMs to learn tasks and generate translations without updating any parameters. Some researchers have further explored supervised fine-tuning for LLMs in translation tasks. Li et al. (2023); Chen et al. (2023); Alves et al. (2023) have improved translations by designing instructions with parallel sentences to fine-tune LLMs. LLMs require extensive training on large data sets for good results, posing challenges for LRLs that consequently result in poor performance (Jiao et al., 2023; Hendy et al., 2023).

To improve translations of LRLs with LLMs, and alleviate the linguistic noise, We dynamically sample and program the auxiliary languages and translation paths during unsupervised training, which broadens linguistic context and enables LLMs to mitigate linguistic noise in translations by UNMT models. In this paper, we present **PrO**bability-driven **M**eta-graph **P**rompter (POMP), which employs a sampling-based dynamic graph incorporating auxiliary pseudo-parallel sentences to prompt LLMs to alleviate the linguistic noise in UNMT methods above and to improve translations in LRLs. Specifically, we employ the **Cross-lingual Transfer NMT Model** (§3.2) (Chen et al., 2022) to generate pseudo-parallel sentences for both source-target and source-auxiliary language pairs. We design a **Language-specific Meta-Graph** (§3.3) to serve as the basis for sampling multiple paths in prompting LLMs. In the meta-graph, vertices denote languages, while edges represent directed translation steps from the source to the pointed language. All paths start with the source and end with the target, through different auxiliary languages. We conduct independent sampling of multiple paths in the meta-graph to build a **Graph-Prompting LLM-based Translator** (§3.4). We propose two kinds of operations: *Generate* and *Aggregate* to prompt LLMs to obtain refined translations for each vertex and the whole path respectively. Furthermore, we design an update strategy **Probabilistic Backward Graph Evolution** (§3.5), in which we update the probability for each auxiliary language considering evaluation scores of translation outcomes, to optimize the prompting graph. After training, we select the prompting graph of the last training instance for inference.

Our contributions are as follows: (1) We propose POMP, a method that constructs a dynamic graph consisting of multiple translation paths with auxiliary languages to prompt LLMs as translators in LRLs. (2) We design a probabilistic backward algorithm for graph evolution that iteratively updates probabilities of auxiliary languages to construct better prompts. (3) We achieve state-of-the-art performance on three LRL pairs in UNMT tasks, in terms of metrics with a high correlation with human evaluation.

## 2 Related Work

### 2.1 Unsupervised Neural Machine Translation on Low-resource Languages

UNMT methods aim to learn an NMT model without parallel data, offering more practical alternatives for LRLs. UNMT methods in recent years are mainly divided into three categories: back-translation, transfer learning, and pivot-based translation.

Back-translation (Sennrich et al., 2016) effectively uses monolingual data to train NMT models by generating synthetic source sentences. Edunov et al. (2018) confirmed its improved performance in both large-scale and low-resource contexts. Lample et al. (2018a); Artetxe et al. (2018) pioneered UNMT using iterative back-translation, mitigating errors of initial NMT models. This method has been effectively applied to LRLs (Chen et al., 2020; Sánchez-Martínez et al., 2020). However, Ed-man et al. (2020) noted that poor initial quality in word embeddings and cross-lingual alignments might reduce translation performance in LRLs.

Transfer learning employs a parent model, usually trained on high-resource languages, to initialize a child model, which is then trained on desired LRLs (Zoph et al., 2016). Chronopoulou et al. (2021) presented a novel meta-learning algorithm to enhance the performance of UNMT models in low-resource domains by leveraging knowledge learned from high-resource domains, using a small amount of training data to quickly adapt to new domains. Moreover, Li et al. (2022) continuously transferred knowledge from a high-resource parent model to a low-resource child model during training, ensuring prediction consistency between them.

Pivot-based translation (Leng et al., 2019) bridges from the source to intermediary languages and then onto the target language, facilitating translation when parallel datasets are insufficient. Kim et al. (2019) applied this concept in transfer learning, using pivot languages and parallel corpora for better source-to-target translation. Currey and Heafield (2019) leveraged monolingual pivot language data to create pseudo-parallel corpora, augmenting data for training. Improper pivot languages chosen in these methods are proven to hinder translation results (Liu et al., 2019). Our approach novelly incorporates a sampling-based trainable graph to select auxiliary languages for efficiently prompting LLMs.

### 2.2 Neural Machine Translation with LLMs

The community has explored various LLMs (Touvron et al., 2023a,b; OpenAI, 2022, 2023), which show promised performance in NMT (Jiao et al., 2023; Hendy et al., 2023; Zhu et al., 2023). Research exploring the translation capabilities of LLMs often involves ICL and fine-tuning methods.

Brown et al. (2020) explored the capabilities of LLMs to learn target tasks with the prompt made up of in-context exemplars and templates. In terms of NMT, Garcia et al. (2023) showed comparable performances of ICL to those large, supervised models. Furthermore, Vilar et al. (2023); Agrawal et al. (2023) evaluated various strategies for selecting translation examples for ICL, emphasizing the importance of example quality.

Moreover, following the idea of instruction tuning (Wei et al., 2022; Chung et al., 2022), (Li et al., 2023) explored enhancing translation by fine-tuning XGLM-7B (Lin et al., 2022), with translation instructions, especially for low-resource languages. (Chen et al., 2023) studied improving LLM's instruction understanding and proposed an instruction dataset collecting negative samples for LLMs' faithfulness.

Research (Alves et al., 2023) has shown that fine-tuning methods perform better than ICL in LRLs translation but at a cost of high computational resources. In contrast, our work integrates two approaches by (1) training a sampling-based graph efficiently compared to billions of parameters, and (2) employing this graph to strategically select auxiliary languages and construct prompts in ICL.

## 3 Methods

### 3.1 Overview

The proposed model, POMP, as shown in Figure 1, comprises four main modules: (1) **Cross-lingual Transfer NMT Model** (§3.2) generates pseudo-parallel sentences for both source-target and source-auxiliary language pairs, where the auxiliary languages are languages of additional sentences in each example of prompts. (2) **Language-specific Meta-Graph** (§3.3) serves as the basis for sampling multiple paths in prompting LLMs. In the meta-graph, vertices denote languages and edges represent directed translation steps from the source to the next connected language. All paths start with the source, pass through multiple auxiliary languages, and end with the target. (3) **Graph-Prompting LLM-based Translator** (§3.4) generate multiple translation paths from the meta-graph by sampling, which we use to prompt LLMs to generate target sentences. (4) **Probabilistic Backward Graph Evolution** (§3.5) updates the probabilities for paths in the meta-graph.

Pseudo-parallel sentences generated by the NMT model (§3.2) are (1) used to calculate probability weights of edges in the meta-graph (§3.3); (2) incorporated as additional sentences in each example of prompts (§3.4) and (3) pseudo-references in evaluation (§3.5). The meta-graph (§3.3) serves as the basis for sampling multiple paths in prompting LLMs (§3.4). The sampling paths involve constructing prompt texts for LLMs (§3.4) to generate target sentences, and these sentences are evaluated against pseudo-references (§3.5). The evaluation scores (§3.5) are back-propagated to update probabilities in the meta-graph (§3.3). Therefore, we achieve a trainable graph incorporating auxiliary languages to prompt LLMs as translators.

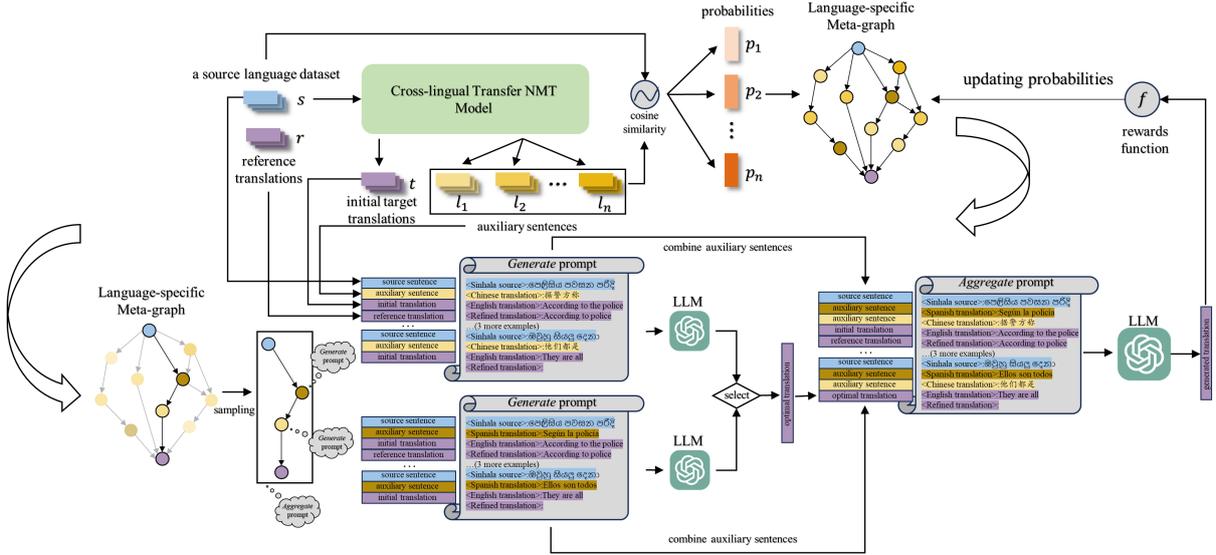

Figure 1: The architecture of POMP. POMP translates source sentences into $n$ auxiliary and target languages using a cross-lingual NMT model, computes average similarity scores to determine auxiliary language probabilities, and constructs a language-specific meta-graph. POMP samples translation paths from the meta-graph and constructs prompt texts for *Generate* and *Aggregate* operations for individual vertices and entire paths. These prompts guide the LLM in generating translations, with the best *Generate* output replacing the initial translation in *Aggregate*. POMP updates auxiliary language probabilities by back-propagating the final *Aggregate* translation score as rewards and then executes the next loop.

### 3.2 Cross-lingual Transfer NMT Model

Following (Chen et al., 2022), we build a cross-lingual transfer NMT model to generate pseudo-parallel sentences for both source-target and source-auxiliary language pairs, where the auxiliary languages are extra languages assisting in prompting LLMs. The model is a Transformer-based NMT model. We first initialize the encoder and embeddings with weights of a multilingual pre-trained XLM-R model (Conneau et al., 2020) and then train with a two-stage training method on six auxiliary language datasets.

In the first stage, we only train the decoder components:

$$\mathcal{L}_{\theta_{\text{dec}}} = \sum_{a_i \in A} \sum_{<\text{x,y}> \in a_i} \log P(\text{y}|\text{x}; \theta) \quad (1)$$

where $A = \{a_1; a_2; \cdots; a_k\}$ is the collection of parallel datasets of $k$ auxiliary languages and English, $<\text{x,y}>$ is a parallel sentence pair, $\theta$ and $\theta_{\text{dec}}$ are the parameter sets of the whole model and its decoder layers respectively.

In the second stage, we jointly optimize all parameters except the embeddings:

$$\mathcal{L}_{\theta_{\text{enc}}, \theta_{\text{dec}}} = \sum_{a_i \in A} \sum_{<\text{x,y}> \in a_i} \log P(\text{y}|\text{x}; \theta) \quad (2)$$

where $\theta_{\text{enc}}$ is the parameter set of the encoder layers.

The first stage aims to preserve the cross-lingual transferability of the encoder when fine-tuning. The second stage aims to further improve the NMT model learning from the training data. Thus, the model transfers its translation ability learned from auxiliary languages in training to LRLs in testing. Experiments show that the NMT model performs well in testing on LRLs. Notably, these LRLs are languages that the model has not been trained on previously.

We utilize the NMT model to generate the initial translations for the LRL source. Then we generate pseudo-parallel sentences of the source-target and source-auxiliary languages. Due to the language bias between high-resource languages in training and LRLs in testing, the initial translations contain linguistic noise that we further aim to mitigate.

### 3.3 Language-specific Meta-Graph

To explore diverse translation paths facilitating cross-linguistic learning of LLMs, we design a language-specific meta-graph, which enables LLMs to learn from rich linguistic data to promote translations in LRLs. We define the meta-graph as $\mathcal{G} = (V, E, W)$, in which vertices ($v \in V$) represent languages within the translation process, while an edge ($e \in E$) signifies a conditional transition from the current vertex to the next vertex, considering all preceding vertices in the path. A path starts from the source, passes through multiple auxiliary languages, and ends with the target. A weight ($w \in W$) assigned to each edge represents the conditional probability of transitioning from the current vertex to the next vertex, given the all previous vertices in the path.

To construct the meta-graph, we apply the following steps: (1) Create two vertices to represent the source language and target language respectively; (2) Connect the source vertex to $m$ different auxiliary vertices. Each auxiliary vertex represents a unique auxiliary language from a set of $m$ available options and is assigned with a unique probability computed by algorithm 1. An edge shows a directed transition from the existing vertex to the new-connected vertex; (3) Further extend directed connections of each auxiliary vertex above to the target vertex and other auxiliary vertices, ensuring that each new connection is a vertex not previously connected in the current path. (4) A path contains contiguous connections between vertices and is complete if it reaches the target vertex, otherwise continue extending by step (3); (5) Weight each directed edge with the joint probability of transitioning from the last vertex to the next vertex. The joint probability is computed by equation 3 considering all probabilities of previous vertices in the path.

In the meta-graph constructed above, we design the algorithm 1 to obtain the probability of an auxiliary language as follows: (1) We first encode pseudo-parallel sentences in §3.2 between the source and the auxiliary language by the NMT model; (2) Then, we compute the average cosine similarity between the pseudo-parallel sentences of the source and the selected auxiliary language. This calculation is based on the representations in the final encoder layer. (3) Finally, we utilize an exponential function to scale the average cosine similarity to the initial probability of the auxiliary language.

---
**Algorithm 1** Calculate Initial Probability
---
1: **Initialization:** sentence pairs set of source $S$ and auxiliary languages $A$; *Encoder* of the translation model;
2: **Input:** pseudo-parallel sentence pairs set in §3.2, $U = \{(s_1, a_1), (s_2, a_2), \ldots, (s_n, t = a_n)\}, s \in S, a \in A$, $n$ is the number of sentence pairs.
3: **for** $(s_i, t_i) \in U, i = 1, 2, \cdots, n$ **do**
4: $\quad \mathbf{v}_{s_i}, \mathbf{v}_{t_i} \leftarrow Encoder(s_i, t_i)$
5: $\quad \text{sim}_i \leftarrow \frac{\mathbf{v}_{s_i} \cdot \mathbf{v}_{t_i}}{\|\mathbf{v}_{s_i}\|\|\mathbf{v}_{t_i}\|}$
6: **end for**
7: $\overline{\text{sim}} \leftarrow \frac{1}{n} \sum_i \text{sim}_i$
8: $p \leftarrow \exp\left(-1 + \overline{\text{sim}}\right)$
9: **Output:** Initial probability of the auxiliary language $p$
---

We execute the algorithm 1 repeatedly for each auxiliary language and get their initial probabilities respectively.

### 3.4 Graph-Prompting LLM-based Translator

To alleviate the linguistic noise in initial translation in §3.2, we build a graph-prompting LLM-based translator that constructs a dynamic graph with multiple auxiliary languages to prompt the LLM to translate. The graph consists of multiple paths that are employed to construct prompt text with auxiliary languages.

Specifically, we employ three operations: *Sample*, *Generate*, and *Aggregate*. We independently *Sample* multiple paths from the language-specific meta-graph. As shown in the table 1, in a path, individual auxiliary language involves a prompt text at a vertex level in the *Generate* operation, and all auxiliary languages involve another prompt text at a path level in the *Aggregate* operation.

### 3.4.1 Sample

To enhance cross-linguistic understanding of LRLs translation in LLMs, we sample multiple translation paths from the meta-graph in §3.3 by the joint probabilities of edge weights in paths. we obtain the joint probability $w$ by:

$$p_{[1,m]} = (\prod_{j=1}^{m} p_j)^{\frac{1}{m}} \tag{3}$$

where $p_j$ is the probability of an auxiliary language and $m$ is the number of auxiliary languages. The joint probability represents the probability of sampling all the auxiliary languages in the path at the same time and will be updated during training.

These sampled paths provide diverse auxiliary languages and their combinations to construct prompt texts in operation *Generate* and *Aggregate*, facilitating a better cross-linguistic understanding of LLMs.

### 3.4.2 Generate

To preserve the fine-grained accuracy of a translation, we execute the operation of *Generate* for each vertex in a sampled path. As specifically shown in table 1, following ICL methods, we construct the prompt with 4-shot examples and 1 query. Each example consists of 4 parts: a random source sentence (e.g.<Sinhala source>), its pseudo-parallel sentence in the auxiliary language at the vertex (e.g.<Spanish translation> **OR** <Chinese translation>), its initial translation (e.g.<English translation>), and its reference translation (e.g.<Refined translation>). The query contains a testing source sentence and its corresponding identical parts except for the reference, which the LLM is prompted to generate.

We conduct the *Generate* operation $N$ times for a testing source sentence through $N$ auxiliary language vertices, resulting in $N$ generated translations. As we acquire pseudo-parallel target sentences of testing source sentences in §3.2, We evaluate each generated translation against its pseudo-parallel target sentence and then retain the best one among them and the initial translation, setting this as the refined translation for the *Aggregate* operation.

### 3.4.3 Aggregate

To promote the coarse-grained cross-linguistic understanding of a translation, we execute the operation of *Aggregate* for the entire sampled path. Similar to *Generate*, the *Aggregate* operation employs a 4-shot example and 1 query format as illustrated in table 1. However, *Aggregate* differs in its approach: (1) it combines pseudo-parallel sentences from all auxiliary languages along the path (e.g.<Spanish translation> **AND** <Chinese translation>), whereas *Generate* uses sentences from a single auxiliary language at each vertex; (2) the query in *Aggregate* includes the refined translation from the *Generate* instead of the initial translation.

The frequency of executions and the utilization of evaluation scores are another two key differences. While *Generate* operates $N$ times to produce $N$ translations, *Aggregate* is conducted once per path, yielding one translation. This translation is evaluated in the same way as *Generate* but its score informs back-propagated rewards to update the probabilities of each auxiliary language in the path. This distinction underlines *Aggregate*'s role in synthesizing insights from the entire path, compared to the vertex-specific approach in *Generate*.

## 3.5 Probabilistic Backward Graph Evolution

To back-propagate evaluation scores in *Aggregate* to update the probabilities of auxiliary languages, we introduce the probabilistic backward graph evolution.

We claim that as for *Aggregate* operation, each auxiliary language in the path contributes to the evaluation score of the generated translation. Therefore, we apportion this score into individual rewards for each auxiliary language as follows.

We define $e_i$ as the *Generate* score at the $i_{th}$ vertex in the path, $E$ as the *Aggregate* score for the whole path, and $d_i$ as the contribution that the auxiliary language at the $i_{th}$ vertex make to $E$. The difference value between $E$

| type of prompt | components | context |
|---|---|---|
| prompt in Generate with Spanish | 1 example | \<Sinhala source\>: පුලිසිය පවසන පරිදි, ඡායාරූප ශිල්පියා හැපුණු වාහනයේයදුරුට අපරාධ චෝදනා එල්ල වීමට ඉඩක් නැත.<br>\<Spanish translation\>: Según las autoridades policiales, el conductor del vehículo que fue llevado por el fotógrafo no tiene ninguna posibilidad de ser acusado de delito.<br>\<English translation\>: According to the police, the driver of the car where the photographer was kidnapped has no chance of being charged with a crime.<br>\<Refined translation\>: According to police, the driver of the vehicle that hit the photographer is unlikely to face criminal charges. |
| | 3 more examples | ... |
| | 1 query | \<Sinhala source\>: ඔවුහු සියල දනො අනතුර සිදුවී තිබූ ස්ථානයෙන් ආපසු දිව ගියහ.<br>\<Spanish translation\>: Todos volvieron de vuelta desde el lugar donde se produjo el accidente.<br>\<English translation\>: They all returned from the location where the accident occurred.<br>\<Refined translation\>: |
| prompt in Generate with Chinese | 1 example | \<Sinhala source\>: පුලිසිය පවසන පරිදි, ඡායාරූප ශිල්පියා හැපුණු වාහනයේයදුරුට අපරාධ චෝදනා එල්ල වීමට ඉඩක් නැත.<br>\<Chinese translation\>: 警方表示，被摄像人乘坐的汽车司机没有可能被指控犯有罪行。<br>\<English translation\>: According to the police, the driver of the car where the photographer was kidnapped has no chance of being charged with a crime.<br>\<Refined translation\>: According to police, the driver of the vehicle that hit the photographer is unlikely to face criminal charges. |
| | 3 more examples | ... |
| | 1 query | \<Sinhala source\>: ඔවුහු සියල දනො අනතුර සිදුවී තිබූ ස්ථානයෙන් ආපසු දිව ගියහ.<br>\<Chinese translation\>: 他们全都从事故发生地跑了回去。<br>\<English translation\>: They all returned from the location where the accident occurred.<br>\<Refined translation\>: |
| prompt in Aggregate with Spanish and Chinese | 1 example | \<Sinhala source\>: පුලිසිය පවසන පරිදි, ඡායාරූප ශිල්පියා හැපුණු වාහනයේයදුරුට අපරාධ චෝදනා එල්ල වීමට ඉඩක් නැත.<br>\<Spanish translation\>: Según las autoridades policiales, el conductor del vehículo que fue llevado por el fotógrafo no tiene ninguna posibilidad de ser acusado de delito.<br>\<Chinese translation\>: 警方表示，被摄像人乘坐的汽车司机没有可能被指控犯有罪行。<br>\<English translation\>: According to the police, the driver of the car where the photographer was kidnapped has no chance of being charged with a crime.<br>\<Refined translation\>: According to police, the driver of the vehicle that hit the photographer is unlikely to face criminal charges. |
| | 3 more examples | ... |
| | 1 query | \<Sinhala source\>: ඔවුහු සියල දනො අනතුර සිදුවී තිබූ ස්ථානයෙන් ආපසු දිව ගියහ.<br>\<Spanish translation\>: Todos volvieron de vuelta desde el lugar donde se produjo el accidente.<br>\<Chinese translation\>: 他们全都从事故发生地跑了回去。<br>\<English translation\>: They all returned from the location where the accident occurred.<br>\<Refined translation\>: |

Table 1: A sample of prompts in the translation path "Sinhala→Spanish→Chinese→English". In this sample. there are three types of prompts. The first two are the prompts in *Generate* at a vertex level, involving Spanish and Chinese pseudo-parallel sentences respectively. The last is the prompt in *Aggregate* at a path level, involving Spanish and Chinese pseudo-parallel sentences together.

and $e_i$ can be estimated by:

$$\begin{cases} E - e_1 = d_2 + d_3 + \cdots + d_m \\ E - e_2 = d_1 + d_3 + \cdots + d_m \\ \vdots \\ E - e_m = d_1 + d_2 + \cdots + d_{m-1} \end{cases} \quad (4)$$

, where $m$ is the number of vertex in the path. Furthermore, we can calculate the individual contribution $d_i$ for each auxiliary language:

$$d_i = \frac{\sum_{i=1}^{m}(E - e_i) - (E - e_i)}{m - 1} \quad (5)$$

To make the training process steady and constrain the probability in $[0, 1]$, we employ a central-symmetry Swish (cs-Swish) function (Ramachandran et al., 2017) to scale the contribution $d_i$ to the reward $r_i$.

$$\text{Sigmoid}(x) = \frac{1}{1 + e^{-x}} \quad (6)$$

$$\text{Swish}(x) = x \cdot \text{Sigmoid}(x) \quad (7)$$

$$\text{cs-Swish}(x) = \begin{cases} -\text{Swish}(-x), & \text{if } x \leq 0 \\ \text{Swish}(x), & \text{if } x > 0 \end{cases} \quad (8)$$

So we can get the reward:

$$r_i = \begin{cases} -d_i \cdot \frac{1}{1+e^{d_i}}, & \text{if } d_i \leq 0 \\ d_i \cdot \frac{1}{1+e^{-d_i}}, & \text{if } d_i > 0 \end{cases} \quad (9)$$

Furthermore, we back-propagate the individual reward $r_i$ to update the probabilities of the $i_{th}$ auxiliary language by:

$$p_i^{\text{new}} = (1 + lr \cdot r_i) \cdot p_i^{\text{old}} \quad (10)$$

, where $p_i^{\text{new}}$ is the updated probability for the auxiliary language at the $i_{th}$ vertex, calculated from the old probability $p_i^{\text{old}}$ and the reward $r_i$, scaled by the learning rate $lr$, which decreases in training.

## 4 Experiments

### 4.1 Languages

Following Chen et al. (2022), we utilize German (De), Spanish (Es), Finish (Fi), Hindi (Hi), Russian (Ru), and Chinese (Zh) as the auxiliary languages, which are high-resource languages from different language families. We select Gujarati (Gu), Kazakh (Kk), and Sinhala (Si) to train their respective prompting graph with our approach and then to test.

### 4.2 Datasets

To construct pseudo-parallel datasets for training, we collect datasets from the OPUS[1] (Tiedemann, 2012, 2016). Specifically, the datasets are from WMT[2] (Gu, Kk) and CCAligned[3] (Si). Then we randomly sample 1000 sentences for each dataset and translate the source side with the NMT model in §3.2. The testing data is from newstest[4] (Gu, Kk) and Flores-200 Testset[5] (Team et al., 2022) (Si), same as our baselines.

### 4.3 Evaluation

Recent work has shown that n-gram metrics like BLEU (Papineni et al., 2002) are suboptimal for evaluating high-quality translations (Kocmi et al., 2021; Freitag et al., 2021). As recommended in Freitag et al. (2022), neural network-based metrics demonstrate a high correlation with human evaluation. Therefore, we adopt BLEURT[6] (Sellam et al., 2020) as our evaluation metric. Specifically, we use *BLEURT-20*[7] model in Pytorch implementation[8] for BLEURT metric.

### 4.4 Baselines

**SixT+** model (Chen et al., 2022) learns from an amount of high-resource language data and inference on LRLs. We use ChatGPT (OpenAI, 2022) as the LLM in **POMP**. As a comparison, we adopt the direct translation (**ChatGPT$_{trans}$**) and the direct refinement (**ChatGPT$_{refine}$**) via 4-shot ICL. Both ChatGPT$_{trans}$ and ChatGPT$_{refine}$ utilize a 4-shot ICL method but with distinct approaches. An example in ChatGPT$_{trans}$ consists of a source sentence and its target translation, and it prompts the LLM to generate the target translation from the source sentence in a query. Conversely, ChatGPT$_{refine}$ includes a source sentence, its initial translation, and a reference translation in its examples, and prompts the LLM to refine the initial translation given the source sentence in a query. Table 2 shows an example for ChatGPT$_{trans}$ and ChatGPT$_{refine}$.

### 4.5 Overall performance

Table 3 shows the results of all methods on the BLEURT metric. Chen et al. (2022) proposed the SixT+ model and achieved previous state-of-the-art translations in the three LRLs. Dabre et al. (2017) claims that linguistic noise in transfer learning impacts translation performances. Therefore, POMP utilizes a dynamic graph to construct prompt text with auxiliary languages to enhance ChatGPT's translations in LRLs. Results show that POMP mitigates linguistic noise to achieve the best performances in LRLs. As a comparison, we conduct another two baselines named ChatGPT$_{trans}$ and ChatGPT$_{refine}$, which employ the same ICL method as POMP to prompt ChatGPT to translate straightforwardly given the source sentence and to refine initial translation without auxiliary languages. Results show that POMP better improves ChatGPT's translations in LRLs than both ChatGPT$_{trans}$ and ChatGPT$_{refine}$.

In ChatGPT$_{trans}$, we conduct straightforward translations in these three LRLS via ChatGPT and it achieves the worst results in all approaches. It proves that LLMs' translations in LRLs are challenging and can be greatly improved by POMP.

In ChatGPT$_{refine}$, we employ ChatGPT to refine an initial translation and results show that ChatGPT captures the idea of refinement and improves a lot than ChatGPT$_{trans}$ based on initial translations, even without auxiliary languages. However, POMP employs multiple auxiliary languages to construct prompts and achieves better results than ChatGPT$_{refine}$.

---

[1] https://opus.nlpl.eu/
[2] https://opus.nlpl.eu/WMT-News-v2019.php
[3] https://opus.nlpl.eu/CCAligned.php
[4] http://data.statmt.org/wmt19/translation-task/test.tgz
[5] https://tinyurl.com/flores200dataset
[6] https://github.com/google-research/bleurt
[7] https://github.com/google-research/bleurt/blob/master/checkpoints.md
[8] https://github.com/lucadiliello/bleurt-pytorch

| type of prompt | components | context |
|---|---|---|
| prompt in ChatGPT$_{\text{trans}}$ | 1 example | <Sinhala source>: පළිසිය පවසන පරිදි, ඡායාරූප ශිල්පියා හැපුණු වාහනයේදුවූවට අපරාධ වරුනෝ එල්ල වීමට ඉඩක් නැත.<br><English translation>: According to the police, the driver of the car where the photographer was kidnapped has no chance of being charged with a crime. |
| | 3 more examples | ... |
| | 1 query | <Sinhala source>: ඔවුහු සියලු දනො අනතුර සිදුවී තිබූ ස්ථානයනේ ආපසු දිව ගියහ.<br><English translation>: |
| prompt in ChatGPT$_{\text{refine}}$ | 1 example | <Sinhala source>: පළිසිය පවසන පරිදි, ඡායාරූප ශිල්පියා හැපුණු වාහනයේදුවූවට අපරාධ වරුනෝ එල්ල වීමට ඉඩක් නැත.<br><English translation>: According to the police, the driver of the car where the photographer was kidnapped has no chance of being charged with a crime.<br><Refined translation>: According to police, the driver of the vehicle that hit the photographer is unlikely to face criminal charges. |
| | 3 more examples | ... |
| | 1 query | <Sinhala source>: ඔවුහු සියලු දනො අනතුර සිදුවී තිබූ ස්ථානයනේ ආපසු දිව ගියහ.<br><English translation>: They all returned from the location where the accident occurred.<br><Refined translation>: |

Table 2: A sample of prompts in the baseline ChatGPT$_{\text{trans}}$ and ChatGPT$_{\text{refine}}$. ChatGPT$_{\text{trans}}$ and ChatGPT$_{\text{refine}}$ both utilize a 4-shot ICL framework with subtle differences. In ChatGPT$_{\text{trans}}$, an example consists of a Sinhala sentence and its target translation, and the LLM is prompted to generate the target translation from a given testing sentence in a query. Conversely, ChatGPT$_{\text{refine}}$'s example includes the same Sinhala sentence, an initial translation, and a reference translation, with a query showing the same testing sentence and initial translation, prompting the LLM to refine this translation.

| Baselines | Gu | Kk | Si |
|---|---|---|---|
| SixT+ | 72.08 | 70.09 | 68.90 |
| ChatGPT$_{\text{trans}}$ | 71.77 | 63.80 | 34.05 |
| ChatGPT$_{\text{refine}}$ | 73.98 | 71.42 | 69.44 |
| POMP | **75.20** | **71.84** | **70.17** |

Table 3: Results of all methods on the BLEURT metric. The best results are in bold. The higher values show better performances.

When horizontally comparing the results of POMP, we find Gu achieves the best while Si is the worst. In terms of relationships among auxiliary languages and testing languages, Gu shares closer linguistic ties with Hi, one of the auxiliary languages, which likely aids in better translation. However, Si has unique characteristics influenced by Dravidian languages, which might not be adequately captured by the auxiliary languages. In terms of the evaluation model in the BLEURT metric, we use the *BLEURT-20* model, which is fine-tuned more on Gu and Kk than Si (Pu et al., 2021). Therefore, the model may score more accurately on Gu and Kk.

## 5 Conclusion

In summary, we propose POMP, a novel unsupervised method enhancing LLM translation capabilities LRLs through a dynamic, sampling-based graph of auxiliary languages. This approach involves constructing a directed acyclic meta-graph per source language, where vertices are auxiliary languages and edges represent probabilities. By dynamically sampling multiple paths through the meta-graph, we prompt LLMs to generate more accurate and coherent translations. We use the BLEURT metric for evaluation, with the results driving the updates of probabilities in the meta-graph. Our experiments across three LRLs demonstrate significant improvements in translation quality, underscoring the effectiveness and potential of our method in UNMT for LRLs.